\pdfoutput=1

\documentclass[11pt]{article}

\usepackage[final]{acl}

\usepackage{times}
\usepackage{latexsym}
\usepackage{graphicx}
\usepackage{adjustbox}
\usepackage{amsmath}

\usepackage{listings}
\usepackage[T1]{fontenc}
\usepackage[utf8]{inputenc}
\usepackage{hwemoji}

\usepackage{hyperref}
\usepackage{fancyvrb}
\usepackage{amssymb}
\usepackage{microtype}

\usepackage{inconsolata}
\usepackage{multirow}
\usepackage{booktabs}
\usepackage{subfigure}

\title{Logical Closed Loop: Uncovering Object Hallucinations in Large Vision-Language Models}

\author{Junfei Wu$^{1,2}$, Qiang Liu$^{1,2}$,  Ding Wang$^{1,2}$, Jinghao Zhang$^{1,2}$, Shu Wu$^{1,2}$\thanks{To whom correspondence should be addressed.}, Liang Wang$^{1,2}$, Tieniu Tan$^{1,2,3}$ \\
        $^1$
        New Laboratory of Pattern Recognition (NLPR), 
        \\ State Key Laboratory of Multimodal Artificial Intelligence Systems, \\ Institute of Automation, Chinese Academy of Sciences \\ 
        $^2$School of Artificial Intelligence, University of Chinese Academy of Sciences \\
        $^3$Nanjing University \\
\texttt{junfei.wu@cripac.ia.ac.cn,}
\texttt{qiang.liu@nlpr.ia.ac.cn,} 
\texttt{wangding2024@ia.ac.cn}\\
\texttt{jinghao.zhang@cripac.ia.ac.cn,}
\texttt{\{shu.wu, wangliang, tnt\}@nlpr.ia.ac.cn}
}

\begin{document}
\maketitle

\newcommand{\themodel}{LogicCheckGPT\xspace}
\begin{abstract}
Object hallucination has been an Achilles' heel which hinders the broader applications of large vision-language models (LVLMs). Object hallucination refers to the phenomenon that the LVLMs claim non-existent objects in the image. To mitigate the object hallucinations, instruction tuning and external model-based detection methods have been proposed, which either require large-scare computational resources or depend on the detection result of external models.
However, there remains an under-explored field to utilize the LVLM itself to alleviate object hallucinations. In this work, we adopt the intuition that the LVLM tends to respond logically consistently for existent objects but inconsistently for hallucinated objects. Therefore, we propose a Logical Closed Loop-based framework for Object Hallucination Detection and Mitigation, namely \textbf{LogicCheckGPT}. 
In specific, we devise logical consistency probing to raise questions with logical correlations, inquiring about attributes from objects and vice versa. Whether their responses can form a logical closed loop serves as an indicator of object hallucination. As a plug-and-play method, it can be seamlessly applied to all existing LVLMs. Comprehensive experiments conducted on three benchmarks across four LVLMs have demonstrated significant improvements brought by our method, indicating its effectiveness and generality\footnote{The source code is publicly available at: \href{https://github.com/Hyperwjf/LogicCheckGPT}{https://github.com/Hyperwjf/LogicCheckGPT}.}.


\end{abstract}

\section{Introduction}

\begin{figure}[t]
   \begin{center}
   \includegraphics[width=0.50\textwidth]{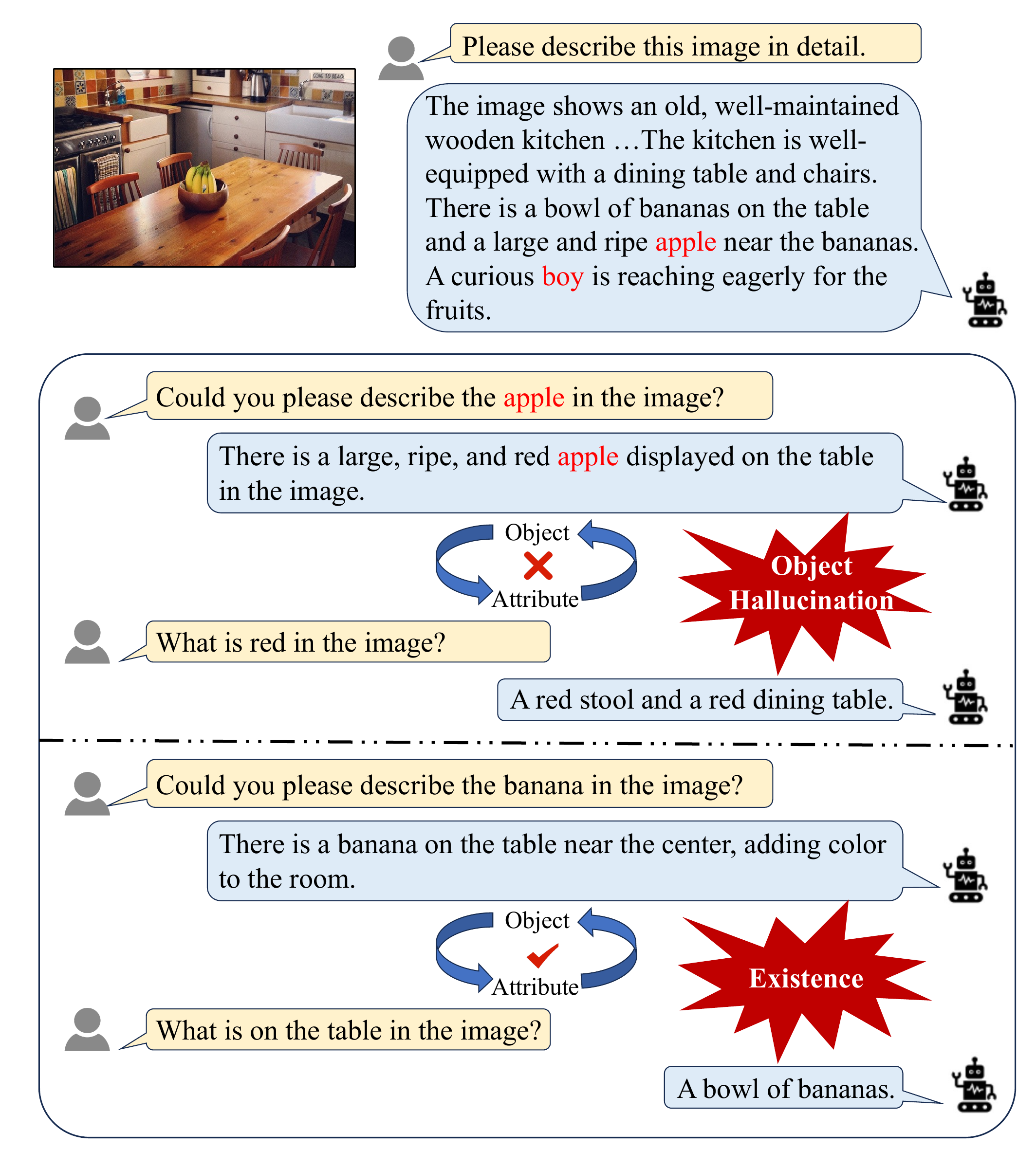}
   \end{center}
   \caption{
    An example of object hallucinations. Hallucinated objects are highlighted in \textcolor{red}{red}. The LVLM shows different logical consistency to hallucinated object ``apple'' and existent object ``banana''. 
   }
   \label{fig:example}
\end{figure}



With the great advancement of large language models (LLMs) \cite{ouyang2022training, touvron2023llama, zhao2023survey}, they have showcased impressive abilities, such as text generation, instruction following. 
Recent studies have been devoted to introduce the powerful capabilities of LLMs to the field of multimodal models. Empowered by LLMs, large vision-language models (LVLMs) \cite{liu2023llava, ye2023mplug, zhu2023minigpt, li2023otter, bai2023qwen,dai2023instructblip} are facilitated to perform strong multimodal understanding and reasoning.

Despite the exciting breakthrough in LVLMs, they all suffer from hallucination issues inevitably, particularly object hallucination.
Object hallucination refers to the phenomenon that the LVLMs  generate inconsistent descriptions of the given image, for example, making up non-existent objects.
Taking Fig. \ref{fig:example} as an example, the LVLM hallucinates several objects which are not existent in the image, including an apple and a boy. This issue hinders more widespread application of LVLMs.
In safety-related scenarios, the consequences of hallucinations would be unbearable.


There have been several efforts made to alleviate hallucinations in LVLMs. 
Generally, these methods can be categorized into three groups. The first as well as the most popular approach \cite{liu2023mitigating, lee2023volcano} mainly resorts to instruction tuning or specific retraining to facilitate LVLMs to generate less hallucinated contents. 
The second approach \cite{yin2023woodpecker, zhou2023analyzing} incorporates external detection models or specific LVLMs to enhance visual understanding, thereby refining the outputs of the original LVLMs. 
The last approach \cite{huang2023opera, leng2023mitigating} investigates the decoding process of LVLMs and devise novel decoding strategies to avoid hallucinations. 
Although these methods have achieved some effectiveness in alleviating hallucinations, there are still some drawbacks: requiring significant computational resources, depending on external models, or necessitating access to the internal parameters of the model.

However, we argue that the logical consistency of LVLM behaviors have the potential to elucidate the underlying hallucination it encapsulates. 
As illustrated in Fig. \ref{fig:example},  we first obtain the attributes ``red'' for ``apple'' and ``on the table'' for ``banana'' by inquiring the LVLM. Subsequently, when we inquire about which object possesses these attributes, the LVLM can correctly respond with ``banana'' but fails to answer for the hallucinated object ``apple''.
It demonstrates that when we pose a series of logically connected questions about a particular object, the LVLM exhibits better logical consistency for existing objects, while its performance tends to degrade for hallucinated objects. 
It is reasonable because the described attributes of hallucinated objects primarily originate from two sources: attributes from other objects in the image, or fabricated attributes absent in the image. Consequently, the model may fail to answer the hallucinated object when we question what possesses these attributes.


Inspired by this observation, we propose a novel and effective framework called Logic Closed Loop for Object Hallucination Detection and Mitigation, namely \textbf{LogicCheckGPT}, which is training-free and only requires language interaction. 
Our aim is to formulate two types of questions in two stages: the first stage involves inquiring attributes based on objects, followed by inquiring objects based on attributes. Whether their responses can form a logical closed loop serves as an indicator of object hallucination.

In specific, according to the different stages of questioning, we divide our framework into 5 steps: 
(1) \textit{Object extraction} extracts objects in the responses of LVLMs.
(2) \textit{Object-to-Attribute inquiring} inquiries into the detailed attributes of the target objects.
(3) \textit{Attribute-to-Object inquiring} further formulates follow-up questions to inquire what object possesses the attributes mentioned in previous answers.
(4) \textit{Logic closed loop check} examines whether the logical relationships from objects to attributes and attributes to objects can form a closed loop.
(5) \textit{Hallucination detection and mitigation} rectifies hallucinated objects if the ratio of closed loops to the total number of questions exceeds a certain threshold.
Our method is a plug-and-play approach that can be applied to various LVLMs without training or relying on external detection models. Furthermore, the question-answer process in natural language enhances its interpretability.

We evaluated the effectiveness of our framework across multiple advanced LVLMs on several benchmarks \cite{Li-hallucination-2023, fu2023mme}, as well as GPT-4v assisted evaluation\cite{liu2023mitigating, yin2023woodpecker}. Our method demonstrates significant improvements across state-of-the-art LVLMs, including a 31.33\%/10.00\% improvement on the POPE dataset for mPLUG-Owl \cite{ye2023mplug}/MiniGPT-4 \cite{zhu2023minigpt}.

Overall, our main contributions can be summarized as follows:
\begin{itemize}
    \item We are the first to adopt the logical closed loop in the context of object hallucination alleviation in LVLMs. 
    \item We propose a novel framework LogicCheckGPT for detecting and mitigating object hallucinations in LVLMs, which is training-free and offers language interaction for user-friendly interpretation.
    \item Comprehensive experiments are conducted to validate the effectiveness of our method, where the results demonstrate the superiority and universality.
\end{itemize}
\section{Related Work}


\subsection{Large Vision-Language Models}
With the surge in the capabilities of large language models (LLMs) \cite{ouyang2022training, zhao2023survey, brown2020language}, there is currently a growing interest in how to integrate the general artificial intelligence of LLMs into the multimodal domains. In consequence, large vision-language models (LVLMs) powered by LLMs are proposed \cite{ye2023mplug, zhu2023minigpt, liu2023llava, li2023otter, dai2305instructblip, bai2023qwen}, which can understand multimodal contents and perform multimodal tasks under instructions. In general, existing LVLMs follow the following paradigm: leveraging a multimodal alignment module to comprehend multimodal inputs, followed by utilizing a LLM to generate responses. Therefore, the training process of LVLMs typically involves modalities alignment pre-training and instruction tuning. Specifically, 
mPLUG-Owl \cite{ye2023mplug} pre-trains the encoder and alignment module, and then finetunes LLaMa \cite{touvron2023llama} by low-rank adaption.
In contrast, LLaVA \cite{liu2023llava} only pre-trains the alignment network and finetunes the alignment network and Vicuna \cite{chiang2023vicuna} on constructed instructions. MiniGPT-4 \cite{zhu2023minigpt} only finetunes the cross-modal alignment network with other modules frozen.

\subsection{Hallucination in LVLMs}
Despite the strong capabilities of these LVLMs, they all grapple with hallucination issues unexpectedly.
To tackle with this issue, 
several benchmarks \cite{fu2023mme, xu2023lvlm, Li-hallucination-2023, lovenia2023negative, jing2023faithscore, chen2024unified} have been proposed to provide detailed evaluations of the hallucination degree exhibited by LVLMs.

Existing hallucination mitigation strategies for LVLMs can be roughly divided into three groups. The first and most widely adopted approach \cite{liu2023mitigating, gunjal2023detecting, lee2023volcano, wang2023vigc} primarily relies on instruction tuning and retraining. 
LRV-Instruction \cite{liu2023mitigating} introduces a comprehensive instruction tuning dataset encompassing positive and negative instructions. 
\cite{wang2023vigc} adopts an iterative instruction generation strategy to improve diversity and accuracy of instructions.
Volcano \cite{lee2023volcano} facilitates the model with the ability to utilize self-feedback to self-revise responses through training. However, these methods heavily depend on the quality of instruction data construction and require substantial computational resources.

The second group of approaches, exemplified by Woodpecker  \cite{yin2023woodpecker} and LURE \cite{zhou2023analyzing}, aim to integrate external detection models or specific LVLMs as revisors to enhance accurate visual understanding, thereby refining base LVLMs' hallucinated generation.
Nevertheless, these approaches rely on external models and fail to explore the intrinsic capabilities of the base model.

The third group of approaches aim to devise decoding strategies to mitigate hallucinations during the process of decoding. 
OPERA \cite{huang2023opera} propose a penalty-based decoding
method along with roll-back strategy to avoid over-trust during decoding. On the other hand, VCD \cite{leng2023mitigating} introduces contrastive decoding to reduce over-reliance on spurious bias and learned priors.
However, obtaining internal states of LVLMs during the decoding process poses a challenge for common users.

Compared to existing approaches, our proposed method is training-free and mitigates hallucinations solely through language interactions. It not only explores the potential of LVLMs to alleviate hallucinations but also offers better interpretability.

\begin{figure*}[t]
   \begin{center}
   \includegraphics[width=0.9\textwidth]{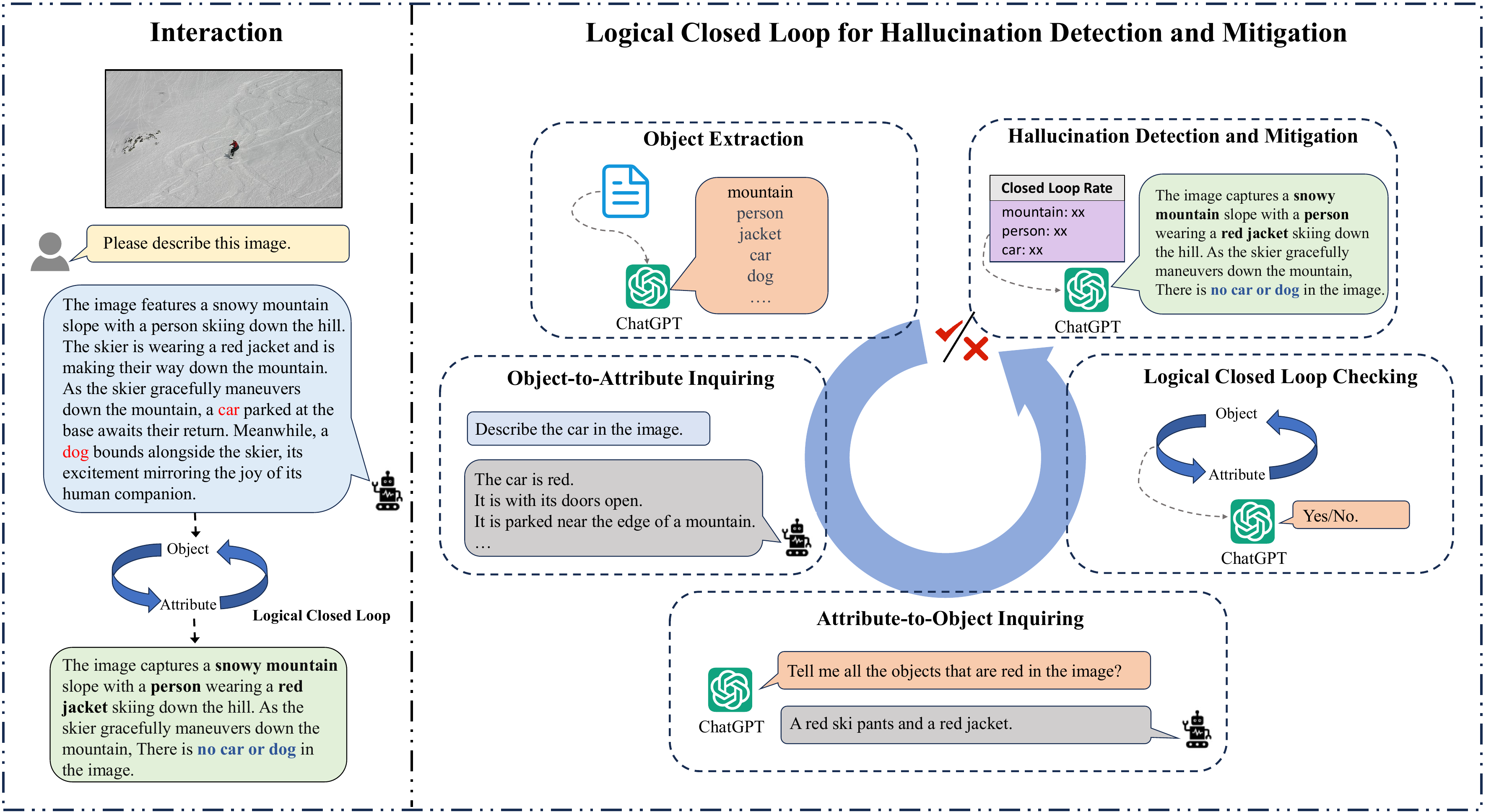}
   \end{center}
   \caption{
    The proposed framework LogicCheckGPT. For LVLM responses to multimodal instructions, LogicCheckGPT employs the following five steps to alleviate object hallucinations: object extraction, object-to-attribute inquiring, attribute-to-object inquiring, logical close loop checking, and hallucination detection and mitigation.
   }
   \label{fig:framework}
\end{figure*}

\subsection{Consistency Checking for Hallucination Detection}

There have also been some works on hallucination detection in LLMs \cite{manakul2023selfcheckgpt, kuhn2022semantic, lin2023generating}, which view the consistency of responses reflects the model's uncertainty.
\cite{kuhn2022semantic} propose semantic entropy to measure the degree of semantic divergence among responses to accommodate semantic equivalence in free-form text.
SelfCheckGPT \cite{manakul2023selfcheckgpt} extend the method to black-box LLMs, eliminating the need for tokens' probability.
\cite{lin2023generating} introduce and compare various uncertainty estimation metrics for black-box LLMs.
In contrast to prior works that focus on consistency among responses to the same question, we propose LogicCheckGPT, a logic consistency-based method that involves logic-related questions and answers. LogicCheckGPT is more capable of delving deeper into the internal uncertainty and the degree of hallucination within the LVLMs.

\section{Method}

In this section, we first introduce the overall framework of LogicCheckGPT, and then elaborate each component. Our framework is shown in Fig. \ref{fig:framework}.

\subsection{Overall}
\label{overall}

To alleviate object hallucinations, we delve into the logical consistency of LVLMs' responses. Specifically, we examine whether the responses demonstrate logical coherence. 
For each object mentioned in a response, we pose two types of sequential questions: one regarding the attributes of the object, and another about which object possesses those attributes.
The logical consistency of the LVLM's responses, i.e., whether they form a closed loop of logical reasoning, serves as an indicator of object hallucination. Here, the ``logical closed loop'' refers to the subsequent object answered being consistent with the initial object.

This process is decomposed
into the following steps: object extraction, object-to-attribute inquiring, attribute-to-object inquiring, logical closed loop checking, and hallucination detection and mitigation.

\subsection{Object Extraction}
\label{object extraction}
To determine the object hallucination, we first need to extract candidate objects from the responses of LVLMs for further querying and checking. For simplicity and versatility, we adopt an LLM to complete each sub-task, including object extraction. Specifically, we employ GPT-3.5 \footnote{https://platform.openai.com/docs/models/gpt-3-5-turbo} as our LLM because of its strong capabilities. The prompt we used can be referred in Appendix \ref{appendix:object extraction}. The extracted objects are represented as \(O = \{o_1, ..., o_i, ..., o_m\}\), where \(m\) is the total number of examinee objects.

\subsection{Object-to-Attribute Inquiring}
\label{first-round question}
As discussed in Section \ref{overall}, the primary step involves constructing object-to-attribute questions to inquire about the attributes of the object. However, it is impractical to enumerate all possible attributes, given their infinite nature. Additionally, it is challenging to create generic attribute question templates applicable to all objects, as different objects typically possess their own specific attributes. For example, attributes related to material for a ``dining table'' differ from those related to clothing for a ``person''.

For better flexibility and adaptability, we prompt the LVLM to provide a detailed description of the object \(o_i\) in free-form text. The template question is formulated as follows: \textit{``Could you please describe the \{object\} in the image?''} We ask the LVLM to respond multiple times. This approach allows us to obtain detailed and specific attribute descriptions from the LVLM regarding the object, thereby facilitating the construction of attribute-to-object questions in subsequent steps. The average number of
extracted attributes of existent or hallucinated objects has also been studied in Section \ref{sec:distribution}.

\subsection{Attribute-to-Object Inquiring}
\label{attribute-to-object prompt}
In the attribute-to-object inquiring stage, our goal is to formulate questions from the attributes to the object, in contrast to Section \ref{first-round question}. After obtaining the attribute descriptions of the object, follow-up questions can be raised based on them. Rather than directly prompting the LLM to formulate questions from the descriptions, we break down this task into two subtasks: attribute extraction and question formulation. This approach avoids potential issues of misleading the LLM to not follow instructions and inadvertently revealing object identity in the questions. It has also been observed that decomposing a task into several simple sub-tasks for LLMs to fulfill yields better performance \cite{wei2022chain, zhao2023survey}.

In specific, we prompt the LLM to extract attributes of the target object \(o_i\) from the description,  wherein the target object is represented as ``The object''.
For instance, extracted attributes such as \textit{``The object is made of wood in the image''}, \textit{``The object is red in color''}. For the examinee object \(o_i\), the extracted attributes are represented as \(A_{i}=\{ a_{i,1},...,a_{i,j},..., a_{i, n_i}\}\), where the number of extracted attributes is denoted as \(n_i\). Detailed prompting instructions can be found in Appendix \ref{appendix:attribute extraction}.

Subsequently, we instruct the LLM to convert these extracted attributes into questions that inquire about what object possesses the specific attributes. 
However, we have noticed that asking questions like \textit{``What is/has \{attribute\} in the image?''}
often yield answers about the most obvious objects with the target attributes, potentially leading to the omission of other less conspicuous yet existent objects.
To deal with this issue,  we frame our questions \(Q_i = \{ q_{i,1},...,q_{i,j},...,q_{i, n_i}\}\) in the format \textit{``Could you tell me all the objects that \{attribute\} in the image?''}. 
The ablation study of this prompt design can be referred in Section \ref{ablation study}.
This format enables the LVLM to comprehensively cover objects that meet the specified attribute constraints.  
Detailed prompting instructions can be found in Appendix \ref{appendix:question formulation}.

\subsection{Logical Closed Loop Checking}

After obtaining the answers \(R_i = \{r_{i,1},...,r_{i,j}, ..., r_{i, n_i}\}\) from the attribute-to-object inquiring stage, we can assess whether each answer forms a logical closed loop, meaning that the object mentioned in the answer is consistent with the examinee object \(o_i\).
In specific, we prompt the LLM to check whether the \(o_i\) is covered in each LVLM's response \(r_{i,j}\). 
The judgement is limited to ``Yes'' and ``No''. 
For the $i$-th examinee object, the judgement of the $j$-th answer is mapped into score $x_{i, j}$ through the mapping \{Yes: $1.0$, No: $0.0$\}.
The prompt is listed in Appendix \ref{appendix:lol}.

\subsection{Hallucination Detection and Mitigation} 

Finally, the logical closed loop rate for each examinee object, defined as the number of logical closed loops divided by the total number of attribute-to-object question-answer pairs \(n_i\), can be formulated as
\begin{align}
    \mathcal{S}(o_i)  = \frac{1}{n_i}\sum_{j}^{n_i} x_{i,j} 
\end{align}
where \(\mathcal{S}(o_i)\) denotes the logical closed loop rate for the \(i\)-th object.



As discussed before, the object is likely to be hallucinated when \(\mathcal{S}(o_i)\) is low. On the other hand, when it's close to 1, it's more likely that the object truly exists. Therefore, \(\mathcal{S}(o_i)\) serves as an indicator of object hallucination. 
Through a valid hallucination threshold \(\lambda\), our method can effectively detect objects that are likely to be hallucinated, i.e., \(\mathcal{S}(o_i)\) is below \(\lambda\).


After identifying hallucinated objects, we guide the LLM to eliminate contents related to hallucinated objects, aiming to minimize hallucinations. The detailed prompt can be found in Appendix \ref{appendix:hallucination mitigation}.

\begin{table*}[t]
    \centering
    \resizebox{0.9\linewidth}{!}{
    \begin{tabular}{l l | c c | c c | c c }
        \toprule
        \multirow{2}{*}{Model} & \multirow{2}{*}{Method} & \multicolumn{2}{c|}{Adversarial} & \multicolumn{2}{c|}{Popular} & \multicolumn{2}{c}{Random} \\
        & & Acc & F1 & Acc & F1 & Acc & F1  \\
         \midrule 
          \multirow{5}{*}{mPLUG-Owl} & vanilla & 50.67 & 66.81 & 51.67 & 67.26 & 55.33 & 68.98 \\
           & LRV-Instruction  & 59.67 & 69.21 & 68.33 & 74.11 & 74.33  & 77.94  \\
           & SelfCheck & 66.67 & 74.09 & 72.00	& 77.29 &  70.66	& 75.82  \\
           & LURE  & 76.33 & 76.72  & 79.67	& 78.75  & 81.33 & 80.95   \\
           &  LogicCheckGPT & \textbf{82.00} &	\textbf{82.23} & \textbf{84.66} & \textbf{84.45} & \textbf{91.00} & \textbf{90.84} \\
           \midrule
          \multirow{5}{*}{MiniGPT-4} & vanilla & 72.67 &	75.88 &	78.33 & 79.87 &	84.33 & 84.59 \\
           &  LRV-Instruction  & 74.00 & 71.11  & 80.33 & 78.70 &  81.67 & 80.97 \\
           & SelfCheck & 73.00 & 72.72 & 76.67 & 75.86 &76.00 & 73.53  \\
           & LURE  & 77.67	& 79.14  & 80.67 & 80.67  &  83.67 & 84.14  \\\
          & LogicCheckGPT & \textbf{82.67} & \textbf{80.59} & \textbf{83.67} & \textbf{81.51} & \textbf{86.67} & \textbf{85.29} \\
          \midrule
          \multirow{4}{*}{LLaVA-1.5} & vanilla & 83.33 & 84.84 &	84.67 & 85.89 & 93.00 & \textbf{93.02}  \\
            &  SelfCheck & 88.67 & 88.27 & 88.67	& 88.59 & 90.33 & 89.53  \\ 
            & LURE  & 85.33	& 86.25 & 87.00 & 87.05 & 89.67 & 89.70  \\
           &  LogicCheckGPT  & \textbf{90.00} & \textbf{89.58} &	\textbf{91.67} &	\textbf{91.40} & \textbf{93.33} & 93.00 \\
           \midrule
           \multirow{4}{*}{QWEN-VL-Chat} & vanilla & 86.67 & 86.67 &	86.33 & 86.37 & 90.67 & 90.28  \\
            & SelfCheck & 87.67 & 87.54 & 87.67 & 87.87 & 91.13 & \textbf{91.45}  \\
            &  LURE  & 87.00 & 87.62 & 87.33 & 87.16 & 88.67 & 88.28  \\
           & LogicCheckGPT  & \textbf{89.00} & \textbf{88.00} &	\textbf{89.67} &	\textbf{88.64} & \textbf{91.33} & 90.71 \\
          \bottomrule
    \end{tabular}
    }
    \caption{The performance comparison between our proposed method LogicCheckGPT and baselines on POPE. The best result is highlighted in boldface.}
    \label{tab:pope}
\end{table*}

\section{Experiment}

\subsection{Experimental Setup}

\subsubsection{Dataset}



\paragraph{POPE} \cite{Li-hallucination-2023} is proposed to provide a detailed evaluation of object hallucination in LVLMs, by querying the models about the presence of specific objects in given images. POPE adopts three sampling settings to construct negative samples: random, popular, and adversarial. The random setting selects non-present objects at random, whereas the popular setting chooses from a list of frequently occurring but absent objects, and the adversarial method selects based on common co-occurrence in contexts despite the absence in the target image.
For each sampling setting, we sample 50 images and 6 questions for each image, with an even distribution of positive and negative samples (50\% - 50\%). 
Accuracy and F1-score are employed as the evaluation metrics.

\paragraph{MME} \cite{fu2023mme} serves as a comprehensive benchmark for evaluating the perecptual and cognitive capabilities of LVLMs across a wide spectrum of tasks. 
For the purpose of this study, we only utilize the \textit{existence} subset to evaluate the phenomenon of object-level hallucination within these models. This approach mirrors the methodology employed in the POPE framework, wherein each subset consists of binary ``Yes-or-No'' questions.  We use accuracy and accuracy+ as metrics, where the former is calculated based on each question, while the latter is based on each image, requiring both questions to be answered correctly.

\paragraph{GPT-4v Assisted Evaluation} To assess the effectiveness of our method for hallucination mitigation in open-ended generation, we adopted the GPT-4v Assisted Evaluation, inspired by \citet{yin2023woodpecker} and \citet{liu2023mitigating}. Our evaluation samples a set of 50 images from the COCO 2014 validation dataset, and asks the model to generate detailed descriptions. Subsequently, we prompt GPT-4v to score original outputs and our outputs in accuracy and relevancy, based on the image and instruction. 
The detailed prompt example is shown in Appendix \ref{appendix:gpt-4v}.

\subsubsection{Baselines}
We selected several widely used open-source LVLMs as backbones to evaluate the effectiveness of our LogicCheckGPT, including mPLUG-Owl (mplug-owl-llama-7b) \citep{ye2023mplug}, LLaVA (llava-1.5-7b) \citep{liu2023llava}, MiniGPT-4 (vicuna-13b) \citep{zhu2023minigpt}, QWEN-VL-Chat \cite{bai2023qwen}. 

We also compare our method with advanced hallucination detection and mitigation methods, including LRV-Instruction \cite{liu2023mitigating}, LURE \cite{zhou2023analyzing} and SelfCheckGPT \cite{manakul2023selfcheckgpt}. 
IRV-Instruction constructs a comprehensive instruction dataset for instruction tuning. However, as they only released the checkpoints of fine-tuned mPLUG-Owl and MiniGPT-4, we report the results of these two models.
LURE trains a LVLM revisor to post-hoc rectify the hallucinated outputs of base models.
SelfCheckGPT employs semantic uncertainty to detect hallucinations. 
We integrate it into our framework to fulfill hallucination mitigation.
In addition, the base LVLMs are referred to as vanilla.




\subsection{Experimental Results}

\begin{table}[t]
  \centering
  \resizebox{0.95\linewidth}{!}{
    \begin{tabular}{llcc}
    \toprule
    Model & Method & Acc & Acc+ \\
    \midrule
    \multirow{5}{*}{mPLUG-Owl} & vanilla & 65.00  & 35.00 \\
    & LRV-Instruction & 83.33 & 66.67 \\
    & SelfCheck & 85.00 & 73.33 \\
    & LURE & 80.00 & 60.00 \\
    & LogicCheckGPT & \textbf{96.67} & \textbf{93.33} \\
    \midrule
    \multirow{5}{*}{MiniGPT-4} & vanilla & 78.33  & 56.67 \\
    & LRV-Instruction & 83.33 & 66.67 \\
    & SelfCheck & 80.00 & 60.00 \\
    & LURE & 85.00 & 70.00 \\
    & LogicCheckGPT & \textbf{86.67} & \textbf{73.33} \\
    \midrule
    \multirow{4}{*}{LLaVA-1.5} & vanilla & 96.67  & 93.33 \\
    & SelfCheck & 96.67 & 93.33 \\
    & LURE  & 93.33  & 86.67 \\
    & LogicCheckGPT & 96.67 & 93.33 \\
    \midrule
    \multirow{4}{*}{QWEN-VL-Chat} & vanilla & 88.33	& 80.00 \\
    & SelfCheck & 93.33 & 86.67 \\
    & LURE & 90.00 & 80.00  \\
    & LogicCheckGPT & \textbf{95.00} & \textbf{90.00} \\
    \bottomrule
    \end{tabular}
    }
  \caption{The performance comparison between our proposed method LogicCheckGPT and baselines on MME Existence subset. The best result on each dataset is highlighted in boldface.}
  \label{tab:mme}
\end{table}%

\begin{table}[t]
  \centering
  \resizebox{\linewidth}{!}{
    \begin{tabular}{llcc}
    \toprule
    Model & Method & Acc & Rel \\
    \midrule
    \multirow{2}{*}{mPLUG-Owl} & vanilla & 3.44  & \textbf{8.78}  \\
    & LogicCheckGPT &  \textbf{4.32} & 8.74 \\
    \midrule
    \multirow{2}{*}{MiniGPT-4} & vanilla & 5.00  & 7.96  \\
    & LogicCheckGPT & \textbf{6.02} & \textbf{8.38}  \\
    \midrule
    \multirow{2}{*}{LLaVA-1.5} & vanilla & 5.22   & 7.24 \\
    & LogicCheckGPT & \textbf{6.50} & \textbf{7.64} \\
   \midrule
    \multirow{2}{*}{QWEN-VL-Chat} & vanilla & 8.36  & 9.96 \\
    & LogicCheckGPT & \textbf{8.58} & 9.96 \\
    \bottomrule
    \end{tabular}
    }
  \caption{The performance of our proposed method LogicCheckGPT over base LVLMs on GPT-4v assisted evaluation. The best result on each dataset is highlighted in boldface.}
  \label{tab:gravie}
\end{table}%

\paragraph{Results on POPE}  
The overall performance of our proposed method LogicCheckGPT on POPE is shown in Table \ref{tab:pope}, from which we have the following observations:


Firstly, our LogicCheckGPT consistently demonstrates significant performance improvements across various LVLMs under different settings.
It can be observed that there is a significant performance decline of LVLMs as we transition from random to popular and adversarial settings, indicating that LVLMs tend to hallucinate non-existent objects that are related to the input image. 
In specific, mPLUG-Owl only achieved accuracies of 50.67\%, 51.66\%, and 55.33\% across the three settings.
For MiniGPT-4, its performance in adversarial settings is over 10\% lower compared to that in random settings. However, with the help of LogicCheckGPT, all LVLMs outperform the vanilla ones significantly under all settings. 
More specifically, mPLUG-Owl equipped with LogicCheckGPT 
achieves an accuracy enhancement exceeding 30\% across three settings.
Despite the already promising performance of LLaVA-1.5, our method still brings a substantial improvement to it, achieving about a 6.67\% increase in accuracy. This demonstrates the effectiveness and robustness of our approach.


Secondly, our proposed LogiCheckGPT outperforms other competitors by a significant margin.
While LRV-Instruction facilitates robust instruction tuning of LVLMs, it remains challenging to completely eliminate potential hallucination issues within the model. 
Due to that SelfCheckGPT relies on semantic consistency in model responses, it is difficult to detect hallucinated objects when the LVLM is overconfident. 
LURE demonstrates significant improvements on models like mPLUG-Owl and MiniGPT-4, but shows only marginal enhancements on more powerful models such as LLaVA-1.5 and QWEN-VL-Chat. We attribute this to the fact that LURE is built upon MiniGPT-4, thus inheriting the limitations of the underlying model.
By contrast, our LogicCheckGPT yields the most substantial improvements across all LVLMs, indicating the superiority of our method.

\begin{figure*}[t] 
    \centering 
    \subfigure[Yes-or-No question.]{ 
        \includegraphics[width=0.48\textwidth]{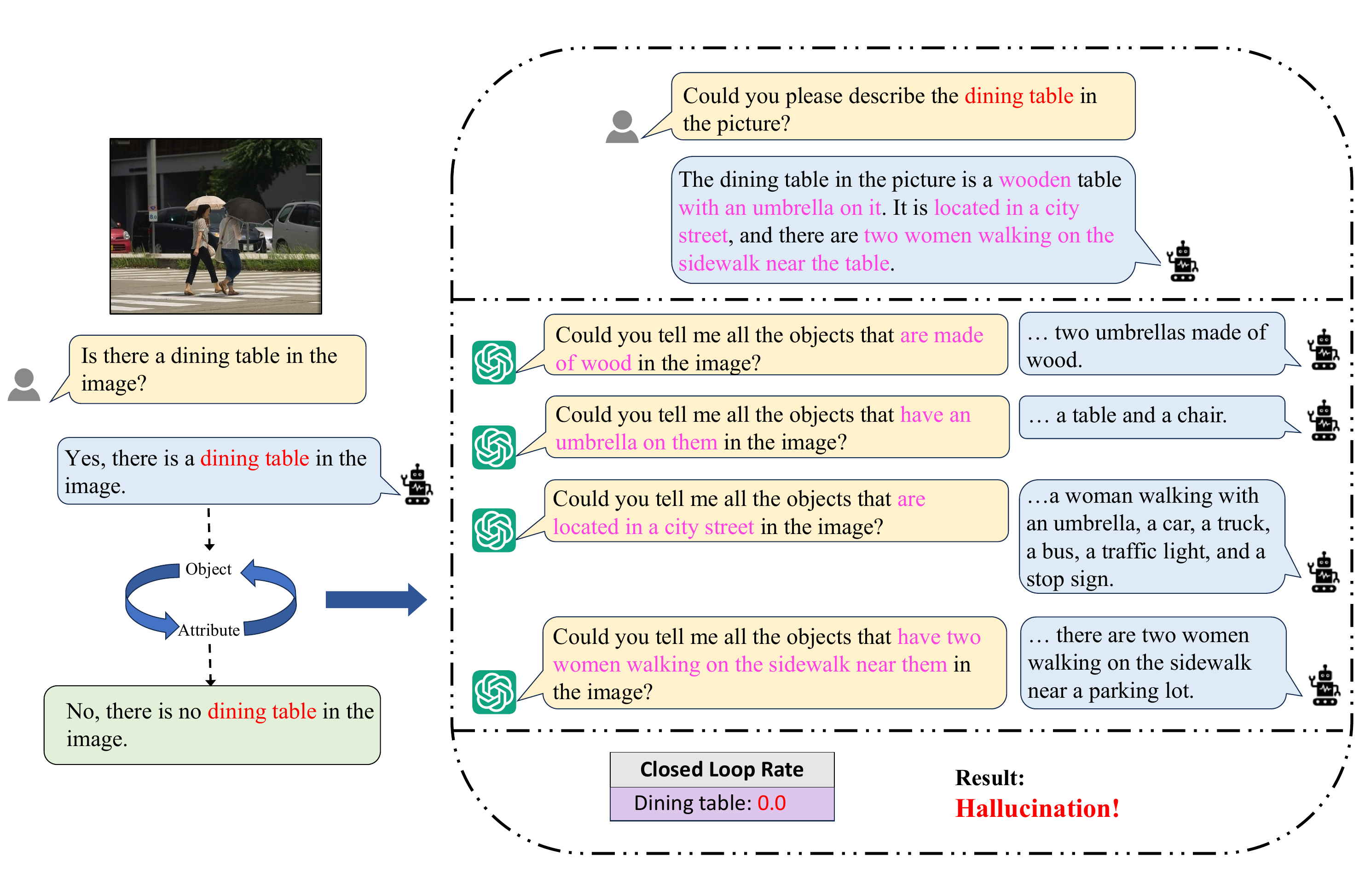} 
    \label{fig:case_1}
    } 
    \subfigure[Open-ended question.]{ 
    \includegraphics[width=0.48\textwidth]{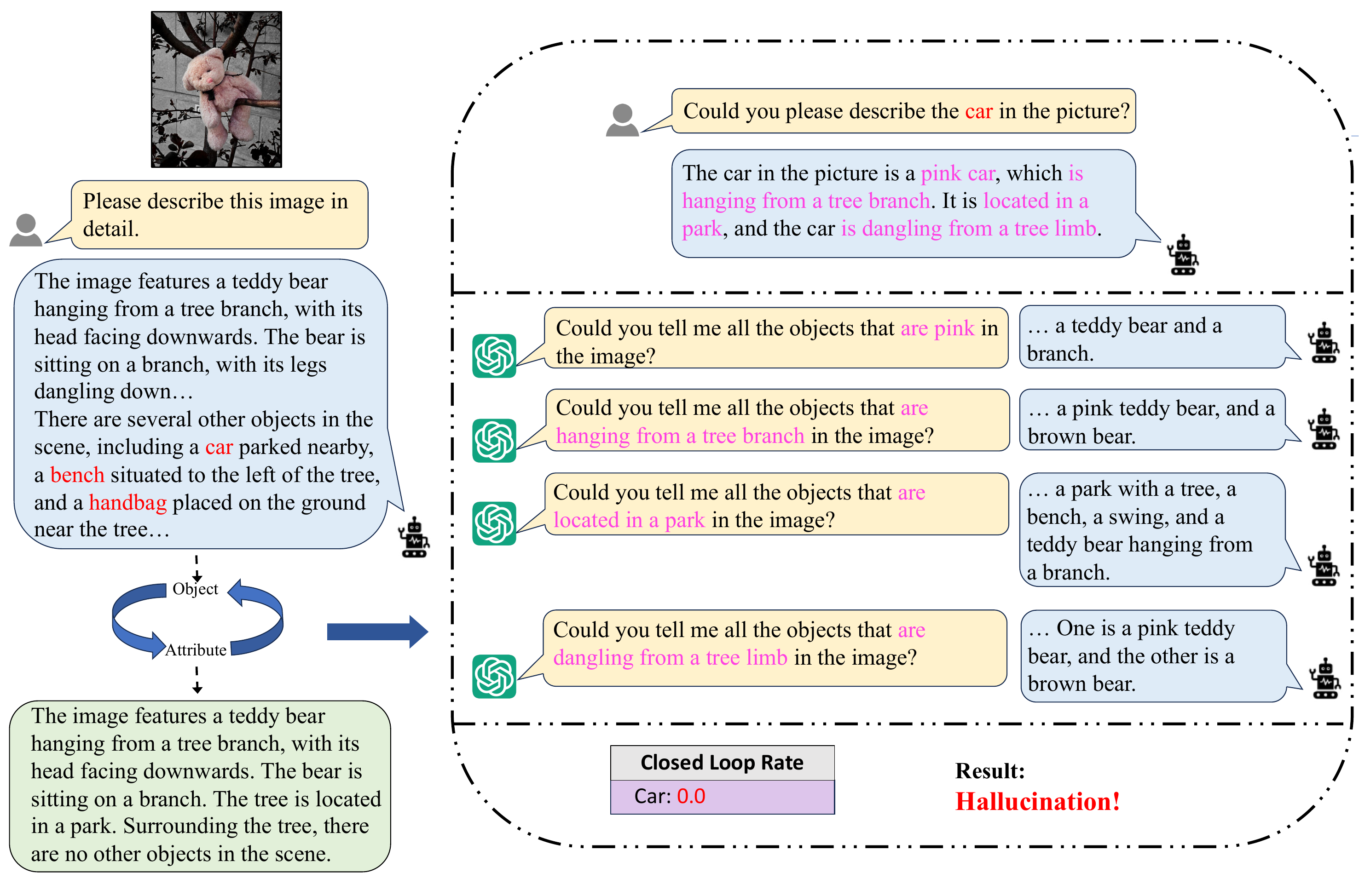} 
    \label{fig:case_2} 
    } 
    \caption{The visualization of two representative examples of our LogicCheckGPT for mPLUG-Owl. The hallucinated objects are highlighted in \textcolor{red}{red} and attributes are highlighted in \textcolor{magenta}{magenta}.} 
\label{fig:case-study} 
\end{figure*}

\paragraph{Results on MME Existence Subset}
As shown in Table \ref{tab:mme}, we also evaluate our method on MME existence subset, which focuses on the object existence hallucination. It can be observed that, although mPLUG-Owl performs worst, our method results in substantial increases in both accuracy and accuracy+, by 31.67\% and 58.33\%, respectively, achieving remarkably high levels of performance. It also demonstrates mPLUG-Owl inherently contains information pertaining to object hallucinations, which can be unearthed by our method. 
In addition, LogicCheckGPT consistently brings significant improvement for MiniGPT-4 and QWEN-VL-Chat, while maintaining the strong performance of LLaVA-1.5.

\paragraph{Results on GPT-4v Assisted Evaluation}
Apart from ``Yes'' or ``No'' questions in POPE and MME, we employed GPT-4v to assist in evaluating our LogicCheckGPT for open-text generation, following previous works \cite{liu2023mitigating, yin2023woodpecker}. The results are summarized in Table \ref{tab:gravie}. 
It can be observed that our method significantly enhances the accuracy of each model, demonstrating the effectiveness of our approach in object hallucination mitigation. In addition, as LogicCheckGPT can remove irrelevant hallucinated information and preserving fluent language structures, it can maintain or even improve relevancy.

\begin{figure}[t]
   \begin{center}
   \includegraphics[width=0.50\textwidth]{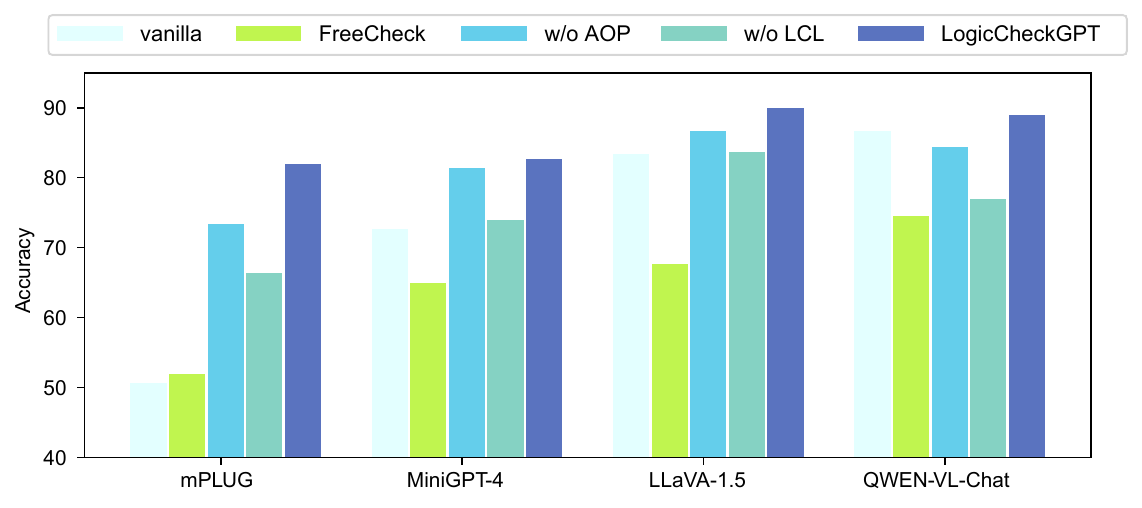}
   \end{center}
   \caption{
    The performance comparison between LogicCheckGPT and several variants, vanilla, FreeCheck, LogicCheckGPT w/o AOP (w/o AOP) and LogicCheckGPT w/o LCL (w/o LCL) across LVLMs on POPE adversarial setting. 
   }
   \label{fig:ablation}
\end{figure}

\subsection{Ablation Study}
\label{ablation study}
We conduct ablation study for several variants, as illustrated in Fig. \ref{fig:ablation}.  
\begin{itemize}
    \item \textit{vanilla} adopts no hallucination mitigation method.
    \item \textit{FreeCheck} employs an LLM (GPT-3.5) to autonomously interrogate LVLMs through multi-turn interactions, similiar to \cite{cohen2023lm}. This iterative process continues until either the LLM reaches a conclusion or the predefined iteration limit (set at 5 here) is reached. The prompt can be referred in Appendix \ref{sec:freecheck}.
    \item \textit{LogicCheckGPT w/o AOP} refers to replacing the Attribute-to-Object prompt in Section \ref{attribute-to-object prompt} by inquire only object instead of covering all objects, e.g. \textit{``What is/has \{attribute\} in the image?''}.
    \item \textit{LogicCheckGPT w/o LCL} employs an LLM to determine the logical consistency directly without Logical Closed Loop rate.
\end{itemize}

We can observe that \textit{FreeCheck} performs worse for most LVLMs, primarily because the LLM fails to raise valuable questions to detect hallucinations. For the existent object, \textit{FreeCheck} often raises questions about other objects that have been mentioned by the LVLMs during the interactions and cannot focus on the examinee object. For the hallucinated object, the questions raised are often simplistic, making it easy for \textit{FreeCheck} to be misled and fail to detect object hallucination.
\textit{LogicCheckGPT w/o AOP} brings significant improvements to several models but still falls below our method, indicating that covering a sufficient number of objects is beneficial.
Though \textit{LogicCheckGPT w/o LCL} falls between vanilla and w/o AOP, demonstrating the effectiveness of calculating logical closed loop rate.
For the powerful QWEN-VL-Chat, which has already achieved impressive performance, only our LogicCheckGPT can enhance its capabilities.

\section{Hyperparameter Analysis}

\begin{figure}[t]
   \begin{center}
   \includegraphics[width=0.50\textwidth]{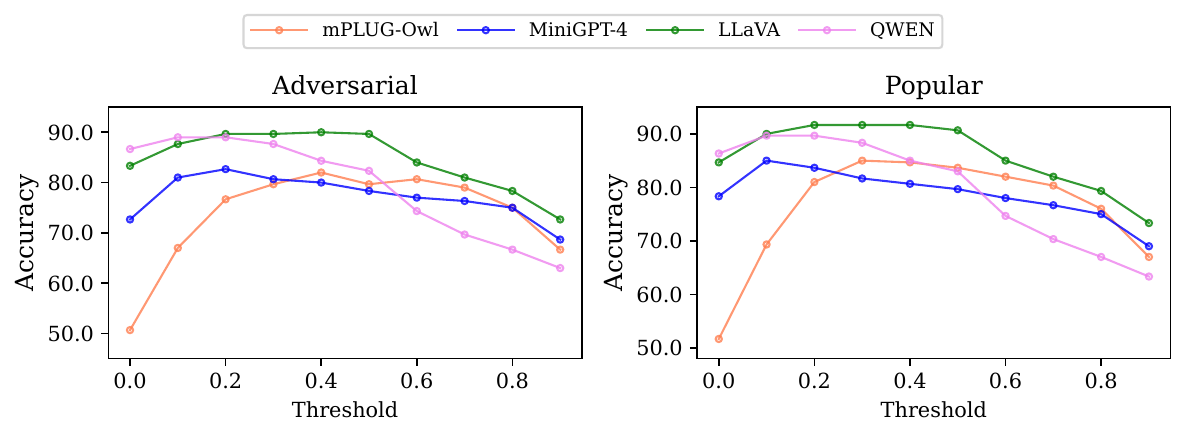}
   \end{center}
   \caption{
    The performance of different threshold $\lambda$.
   }
   \label{fig:threshold}
\end{figure}


In this section, we conduct experiments on POPE under adversarial and popular setting to analyze the performance fluctuation of LogicCheckGPT with different values of key hyperparameters.

\subsection{Logical closed loop threshold $\lambda$}

We test our method with values of threshold $\lambda$ ranging from 0.0 to 0.9 as shown in Figure \ref{fig:threshold} and have the following observations:

There is a significant increase when threshold $\lambda$ is increasing from 0.0 initially for all models with LogicCheckGPT under both settings. It indicates that our method can significantly distinguish existent objects and hallucinated objects by valid threshold, as models tend to provide logically consistent responses for existent objects.

Subsequently, these these models reach their respective performance peaks. For instance, the performance of mPLUG-Owl achieves the best when $\lambda$ is 0.4 under adversarial setting, and the best $\lambda$ for LLaVA is also 0.4. For MiniGPT-4, the $\lambda$ at which it reached its performance peak varied across different settings, 0.2 and 0.1 respectively. With the continued increase of the threshold, a noticeable decrease in performance can be observed. This is reasonable, as it leads to the misclassification of a large number of existent objects as non-existent. 

\subsection{The number of attributes $n$}
\label{sec:num_att}

\begin{figure}[t]
   \begin{center}
   \includegraphics[width=0.50\textwidth]{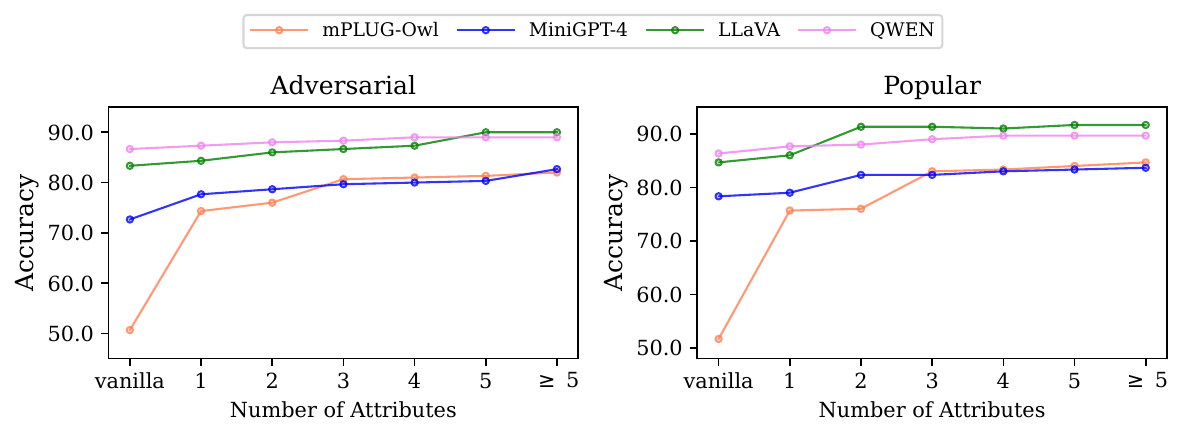}
   \end{center}
   \caption{
    The performance of different number of attributes $n$.
   }
   \label{fig:attribute}
\end{figure}

The number of attributes $n$ obtained during the object-to-attribute stage determines the number of questions that can be generated from these attributes. We report the performance when $n = 1, 2, 3, 4, 5$ and $\ge 5$ (See Figure \ref{fig:attribute}) and summarize the observations as follows:

It can be observed that there is an obvious performance improvement when the number of attributes increases across the four LVLMs. This improvement is reasonable as it allows for a greater range of inquiries to verify the presence of the object. As \(n\) exceeds 5, the improvement becomes marginal.

\subsection{Qualitative Examples}

In this section, 
we have selected two representative examples for mPLUG-Owl covering distinct types of questions, including a binary question, ``Is there a dining table in the image?'' and an open-ended query, ``Please describe this image in detail.'', as illustrated in Fig \ref{fig:case-study}. In the first case \ref{fig:case_1}, the model answers ``yes'' to the binary question, while our method detects that dining table is hallucinated and corrects the output. As for open-ended text generation, the models tends to hallucinate nonexistent objects as the length of the generated sequence increases. However, our method is capable of individually verifying the existence of objects, thereby mitigating hallucinations. More qualitative analysis can be referred in Section \ref{sec:qualtative_analysis}.
\section{Conclusion}

We propose a novel logical closed loop-based framework LogicCheckGPT for object hallucination mitigation in LVLMs.
Our motivation stems from the observation that LVLMs often exhibit logically inconsistent responses to hallucinated objects. Therefore, we devise logic consistency probing, which involves asking questions with logical correlations, such as inquiring about attributes from objects and vice versa. Specifically, we break down this process into several steps: object extraction, object-to-attribute inquiring, attribute-to-object inquiring, logical closed-loop checking, and hallucination detection and mitigation.
Comprehensive experiments conducted on several benchmarks  demonstrate the superiority of our framework.

\section*{Limitations}

In this work, we propose a logical closed loop-based framework for object hallucination mitigation. However, our method still has the following two limitations. 
Firstly, adopting our framework inevitably incurs costs, as we rely on the GPT-3.5 API.
Secondly, our work only focuses on addressing object hallucinations.  There are also other types of hallucinations, including attribute hallucinations and knowledge hallucinations. 
Therefore, extending our framework to encompass a broader range of hallucination mitigation represents our future directions.

\section*{Acknowledgements}
This work is supported by National Natural Science Foundation of China (62372454,62141608, 62236010).


\bibliography{custom}

\appendix
\label{sec:appendix}

\section{Implementation Details}
\label{Implementation Details}

In this work, LogicCheckGPT was constructed utilizing the PyTorch framework \cite{paszke2019pytorch}, incorporating capabilities from HuggingFace's Transformers library \cite{wolf2019huggingface}. 
The large language model (LLM) we adopted is GPT-3.5-turbo, to help fulfill each subtask of hallucination mitigation.
All base LVLMs and hallucination mitigation methods are re-implemented according to their literature. We maintain the default hyperparameter settings for all backbone LVLMs and baselines. 
The experiments were conducted using an NVIDIA A100 GPU and an AMD EPYC 7763 CPU.
The hallucination threshold \(\lambda\) is searched within \([0.0, 0.9]\). The $\lambda$ is set 0.4,0.2,0.4 and 0.2 for mPLUG-Owl, MiniGPT-4, LLaVA-1.5, and QWEN-VL-Chat respectively.  The number of attributes $n$ is set to at least 5. Specifically, in the object-to-attribute inquiring stage, we keep sampling responses from LVLMs until we have obtained at least 5 factored attributes or until we have sampled a maximum of 3 responses.

\begin{table*}[t]
    \centering
    \resizebox{0.9\linewidth}{!}{
    \begin{tabular}{l l | c c | c c | c c }
        \toprule
        \multirow{2}{*}{Model} & \multirow{2}{*}{Method} & \multicolumn{2}{c|}{Adversarial} & \multicolumn{2}{c|}{Popular} & \multicolumn{2}{c}{Random} \\
        & & Acc & F1 & Acc & F1 & Acc & F1  \\
         \midrule 
          \multirow{3}{*}{mPLUG-Owl} & vanilla & 50.67 & 66.81 & 51.66 & 67.26 & 55.33 & 68.98 \\
          &  LogicCheckGPT (Vicuna) & 80.67	& 81.98  & 83.33 &84.17	& 89.00	& 88.96 \\
           &  LogicCheckGPT (GPT) & \textbf{82.00} &	\textbf{82.23} & \textbf{84.66} & \textbf{84.45} & \textbf{91.00} & \textbf{90.84} \\
           \midrule
          \multirow{3}{*}{MiniGPT-4} & vanilla & 72.67 &	75.88 &	78.33 & 79.87 &	84.33 & 84.59 \\
          &  LogicCheckGPT (Vicuna) & 81.00 & 80.27	& 86.00	& 84.78  & \textbf{87.67}	& \textbf{86.34} \\
          & LogicCheckGPT (GPT) & \textbf{82.67} & \textbf{80.59} & \textbf{83.67} & \textbf{81.51} & 86.67 & 85.29 \\
          \midrule
          \multirow{3}{*}{LLaVA-1.5} & vanilla & 83.33 & 84.84 &	84.67 & 85.89 & 93.00 & \textbf{93.02 } \\
            &  LogicCheckGPT (Vicuna) & 87.33 & 88.05 & 87.00 & 87.77 & 93.00	& 92.63 \\
           &  LogicCheckGPT (GPT) & \textbf{90.00} & \textbf{89.58} &	\textbf{91.67} &	\textbf{91.40} & \textbf{93.33} & 93.00 \\
           \midrule
           \multirow{3}{*}{QWEN-VL-Chat} & vanilla & 86.67 & 86.67 &	86.33 & 86.37 & 90.67 & 90.28  \\
           &  LogicCheckGPT (Vicuna) & 87.67 & 87.28 & 88.33 & 87.88 & 90.33 & 89.89 \\
           & LogicCheckGPT (GPT)  & \textbf{89.00} & \textbf{88.00} &	\textbf{89.67} &	\textbf{88.64} & \textbf{91.33} & 90.71 \\
          \bottomrule
    \end{tabular}
    }
    \caption{The performance comparison between LogicCheckGPT implemented by Vicuna and GPT on POPE.}
    \label{tab:variant_vicuna}
\end{table*}

\section{The Choice of LLM for LogicCheckGPT}

To investigate the flexibility of our framework, we have conducted experiments by replacing GPT-3.5 with Vicuna-13b-v1.5\footnote{https://huggingface.co/lmsys/vicuna-13b-v1.5} \cite{chiang2023vicuna} . The results of the Vicuna variant and GPT variant on POPE are summarized in Table \ref{tab:variant_vicuna}.

It can be observed that LogicCheckGPT implemented by Vicuna exhibits strong performance across the four LVLMs. Though the Vicuna variant performs slightly worse than the GPT variant, it still demonstrates a strong capability to detect and mitigate object hallucinations within LVLMs. We also delve into the failure cases of the Vicuna variant and observe that this was due to errors occurring in object extraction and logical closed loop checking. These shortcomings primarily stem from inherent limitations within Vicuna itself and the design of prompts. This underscores the efficacy of our approach and the potential for substituting GPT with alternatives at each stage. As we have unified the framework and divided the whole process into several simple steps, including object extraction, attribute extraction, and rephrasing, we retain the flexibility to substitute GPT with alternative methods tailored to each step, potentially enhancing both efficiency and accuracy.

\section{The distribution of the number of extracted attributes}
\label{sec:distribution}

\begin{table}[t]
  \centering
  \resizebox{0.95\linewidth}{!}{
    \begin{tabular}{lcc}
    \toprule
    Model & \# Existent Objects & \# Hallucinated Objects \\
    \midrule
    
    mPLUG-Owl & \(3.966\pm1.3427\)  & \(3.510\pm1.1783\) \\
    MiniGPT-4 & \(6.576\pm3.6195\) & \(6.322\pm4.4117\) \\
    LLaVA-1.5 & \(5.414\pm1.5077\) & \(5.375\pm1.2747\) \\
    QWEN-VL-Chat & \(4.353 \pm 1.6652 \) & $3.850 \pm 1.1821$ \\
    \bottomrule
    \end{tabular}
    }
  \caption{The average number of attributes extracted in responses when the LVLM is prompted to describe existent or hallucinated objects. The symbol “\#” denotes “the number of”.}
  \label{tab:num_att}
\end{table}%

To investigate whether there is a difference in the distribution of the number of extracted attributes between existent and hallucinated objects, we analyzed the number of attributes extracted from each description for both types of objects across 300 POPE adversarial instances, as shown in Table \ref{tab:num_att}. We have the following observations.

There is a greater abundance of described attributes and a wider variance of existent objects among existent objects compared to hallucinated objects across the four LVLMs. This phenomenon can be attributed to the LVLMs could provide more detailed descriptions of existent objects based on the image while lacking a solid foundation for hallucinated objects. Notably, the MiniGPT-4 tends to describe more attributes for both existent and hallucinated objects, which aligns with our observation that responses generated by MiniGPT-4 are lengthier. However, as the numbers of attributes for both objects are close, it is still difficult to distinguish them only by the attribute count.

In our work, the number of attributes obtained during the object-to-attribute stage is set to at least 5. Specifically, in the object-to-attribute inquiring stage, we keep sampling responses from LVLMs until we have obtained at least 5 factored attributes or until we have sampled a maximum of 3 responses. This approach aims to gather a sufficient number of attributes for the Logical Closed Loop checking process, without being hindered by limitations in attribute quantity. It has been validated in Section \ref{sec:num_att} that as the number of attributes exceeds 5, the improvement becomes marginal.

\section{Prompts}
\label{prompt}

\subsection{Object Extraction}
\label{appendix:object extraction}
\begin{figure*}[t]
   \begin{center}
   \includegraphics[width=1.0\textwidth]{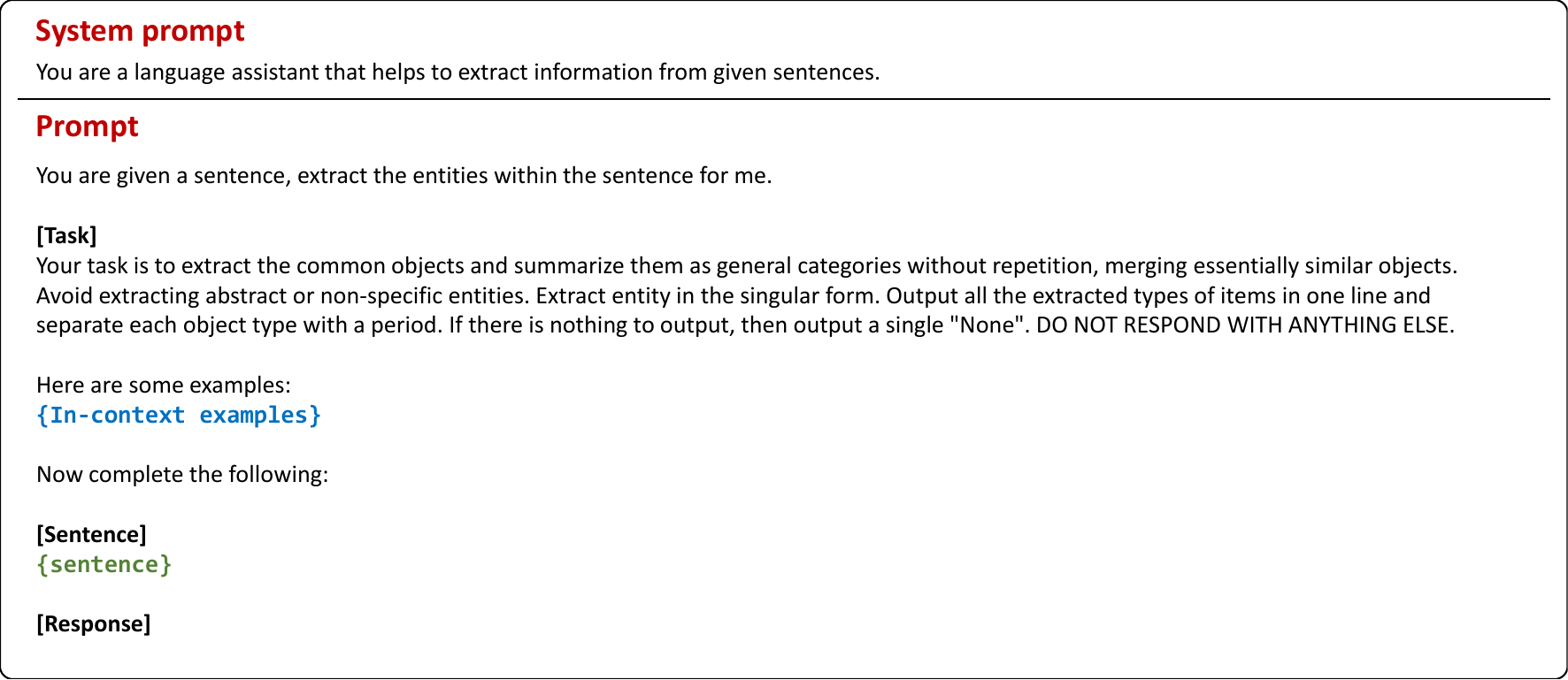}
   \end{center}
   \caption{
     Prompt template of object extraction. 
   }
   \label{fig:app_1}
\end{figure*}

The prompt for object extraction is illustrated in Fig. \ref{fig:app_1}.

\subsection{Attribute-to-Object Inquiring}
\subsubsection{Attribute Extraction}
\label{appendix:attribute extraction}
\begin{figure*}[t]
   \begin{center}
   \includegraphics[width=1.0\textwidth]{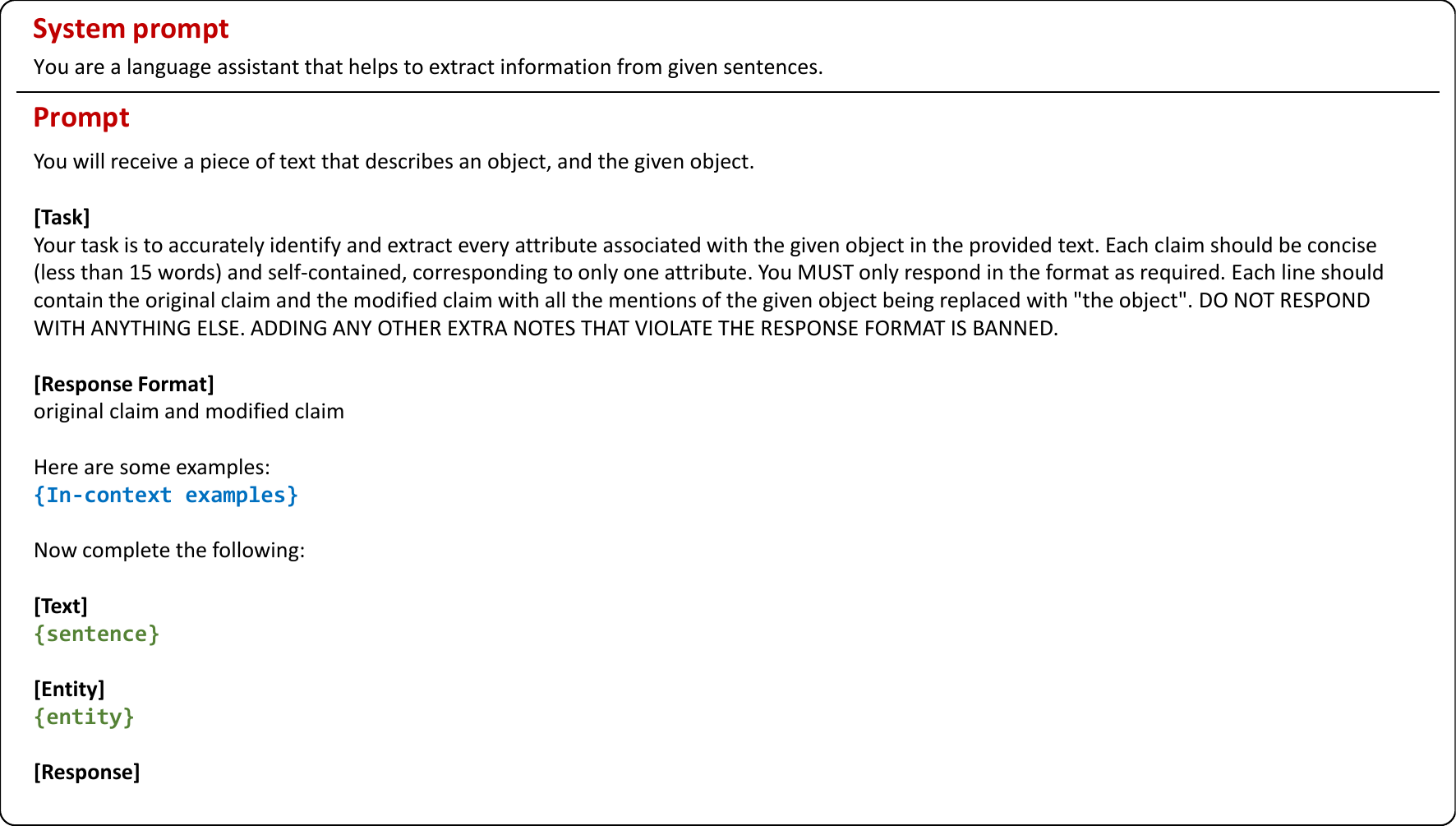}
   \end{center}
   \caption{
     Prompt template of attribute-to-object question (1).
   }
   \label{fig:app_2}
\end{figure*}

The prompt for attribute extraction in attribute-to-object inquiring is illustrated in Fig. \ref{fig:app_2}.

\subsubsection{Question Formulation}
\label{appendix:question formulation}
\begin{figure*}[t]
   \begin{center}
   \includegraphics[width=1.0\textwidth]{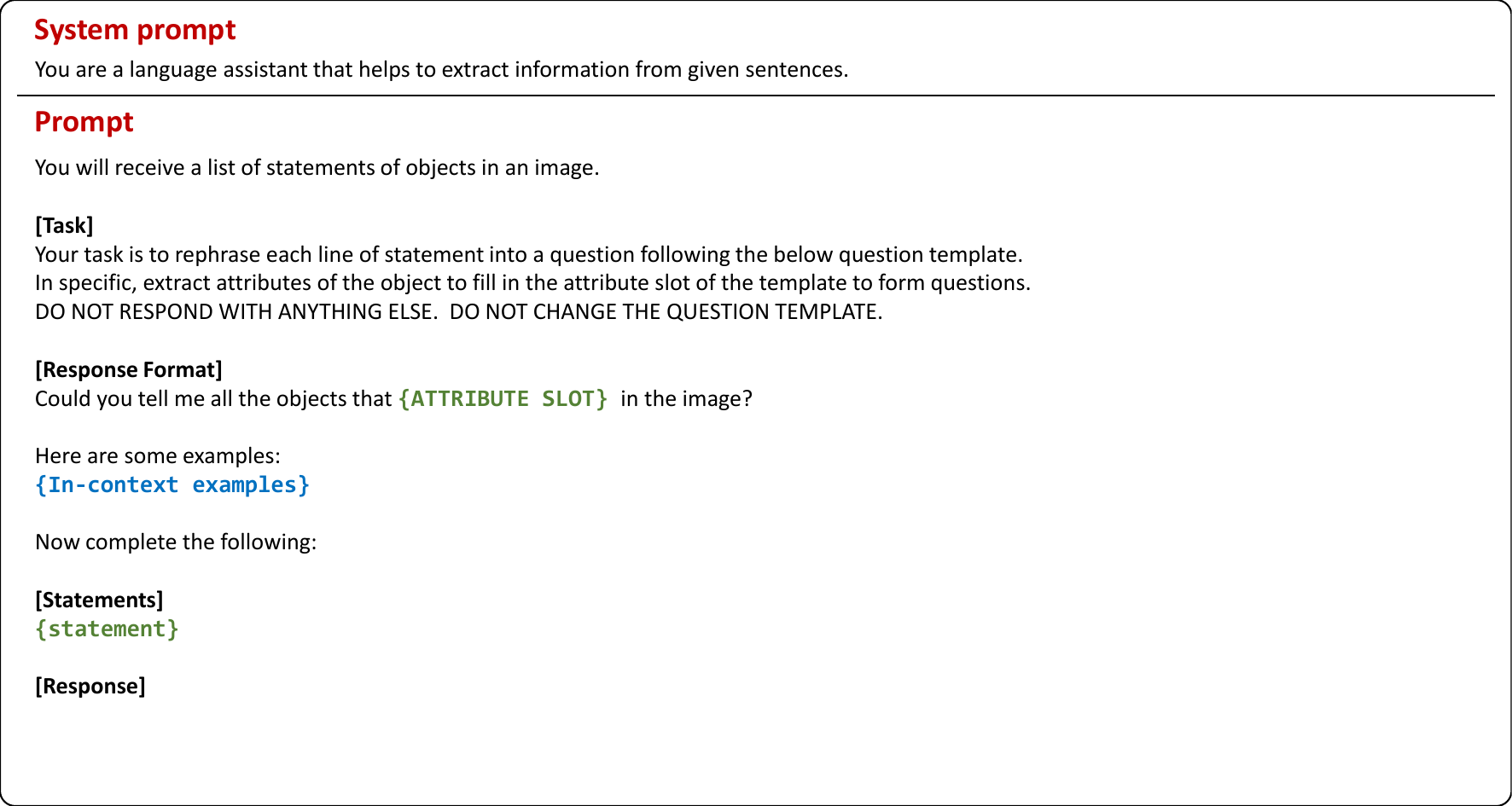}
   \end{center}
   \caption{
     Prompt template of attribute-to-object question (2).
   }
   \label{fig:app_3}
\end{figure*}

The prompt for question formulation in attribute-to-object inquiring is illustrated in Fig. \ref{fig:app_3}.

\subsection{Logical Closed Loop Checking}
\label{appendix:lol}
\begin{figure*}[t]
   \begin{center}
   \includegraphics[width=1.0\textwidth]{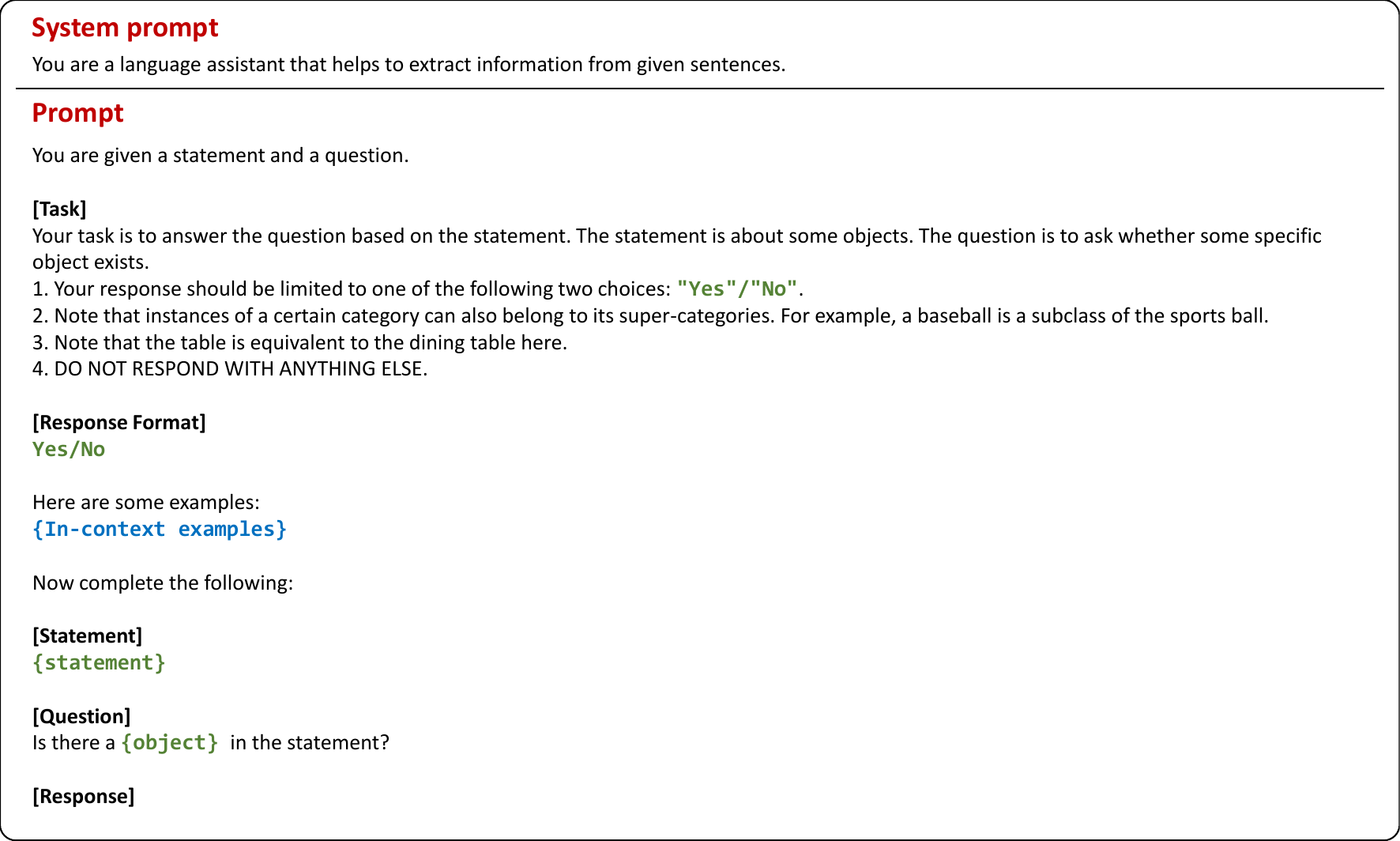}
   \end{center}
   \caption{
     Prompt template of logic closed loop check.
   }
   \label{fig:app_4}
\end{figure*}

The prompt for logical closed loop checking is illustrated in Fig. \ref{fig:app_4}.

\subsection{Hallucination Detection and Mitigation}
\label{appendix:hallucination mitigation}
\begin{figure*}[t]
   \begin{center}
   \includegraphics[width=1.0\textwidth]{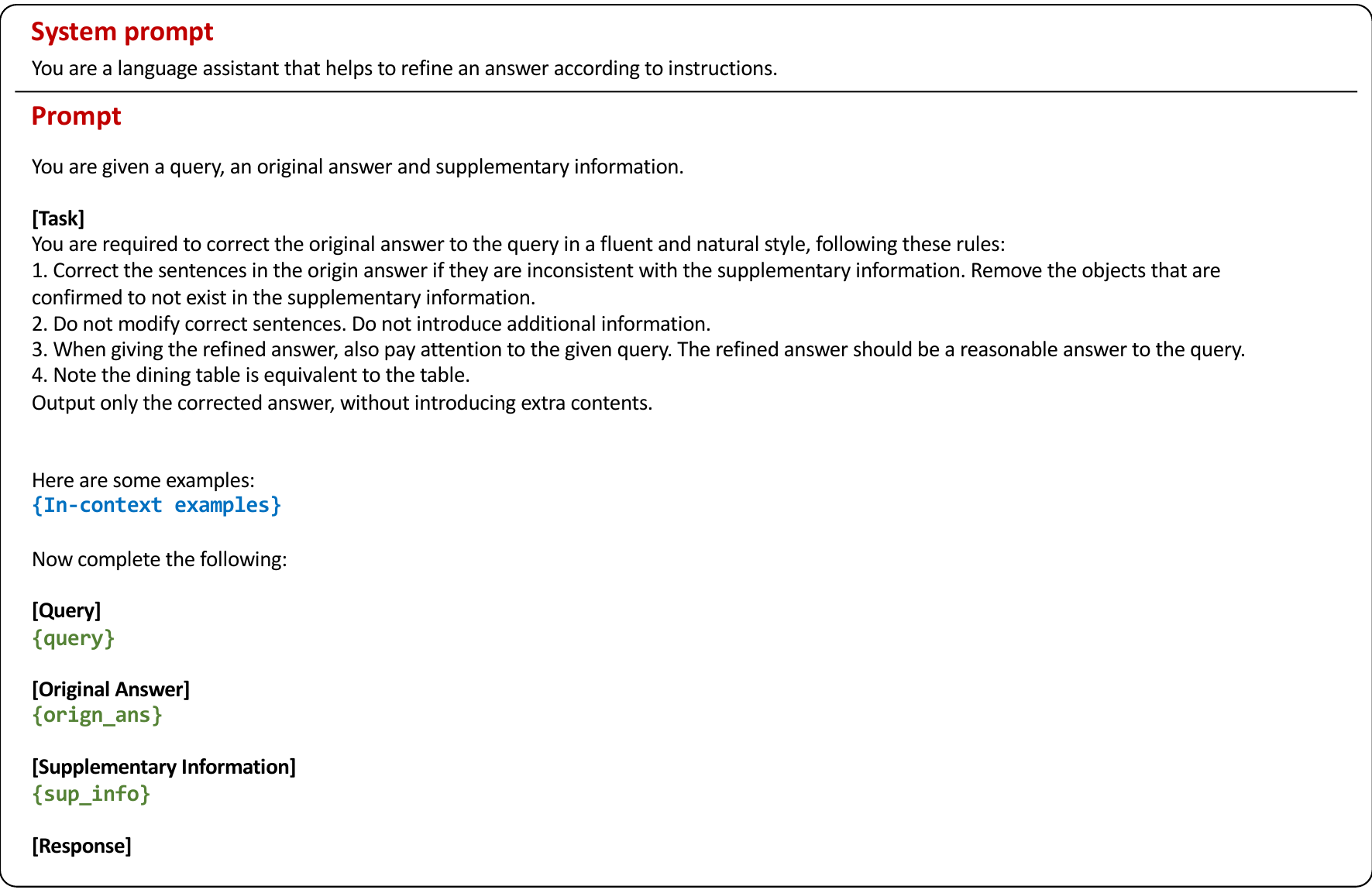}
   \end{center}
   \caption{
     Prompt template of refinement.
   }
   \label{fig:app_5}
\end{figure*}

The prompt for hallucination dection and mitigation is illustrated in Fig. \ref{fig:app_5}.

\subsection{GPT-4V Assisted Evaluation}
\label{appendix:gpt-4v}
\begin{figure*}[t]
   \begin{center}
   \includegraphics[width=1.0\textwidth]{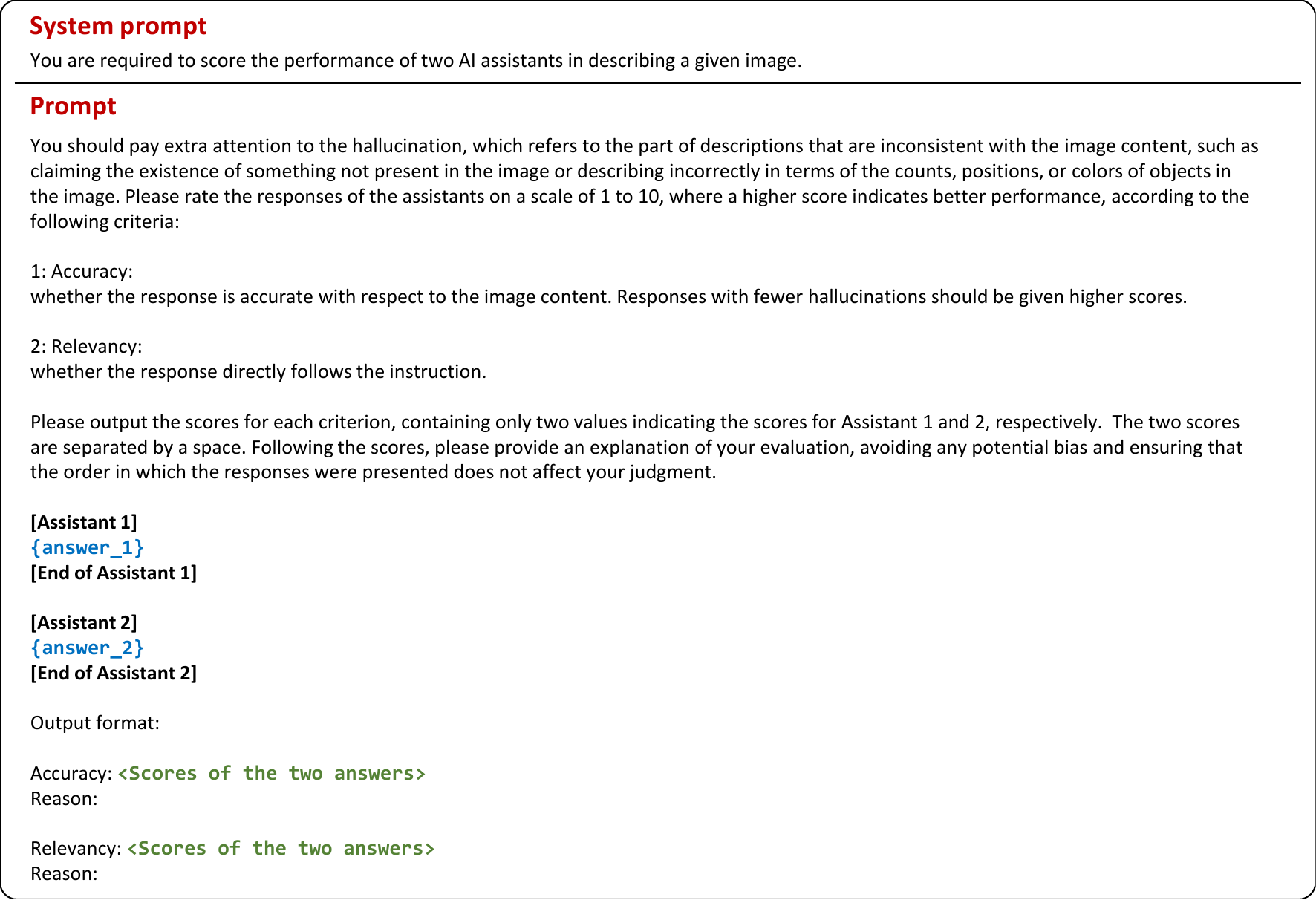}
   \end{center}
   \caption{
     Prompt template of GPT-4v Assisted Evaluation.
   }
   \label{fig:gavie}
\end{figure*}

The prompt for hallucination dection and mitigation is illustrated in Fig. \ref{fig:gavie}.

\section{FreeCheck Prompts}
\label{sec:freecheck}
The prompts for the ablation variant \textit{FreeCheck} in different stages is illustrated in Table \ref{tab:freecheck}.

\begin{table*}[t]
    \centering
    \resizebox{1.0\linewidth}{!}{
    \begin{tabular}{l l }
        \toprule
        \textbf{Stage}  & \textbf{Prompt(s)}  \\
        \hline
        (1) System Prompt
        & You are a language assistant that helps to answer the question according to instructions \\
        \hline
        (2) Setup & Your goal is to validate a description of an image that declares that an object exists in the image. \\
        & You have the opportunity to ask multiple questions to verify the authenticity.\\
        & The claim is: \{claim\}.\\
        & Please start with several questions to ask attributes of the object.\\
        \hline
        (3) Follow-Up Questions & (i) Do you have any follow-up questions? Please only answer with Yes or No. \\
        & (ii) What are the follow-up questions to the claim: \{claim\}?
        \\
        \hline
        (4) Decision & You have ended the question session due to limited rounds. \\
        & Based on the interviewee's answers to your questions, what is your conclusion about the question: \{question\}?  \\
        & Please answer yes/no in the first line and then output explanation.\\
        \bottomrule
    \end{tabular}
    }
    \caption{Prompts used for the ablation variant \textit{FreeCheck} in different stages.}
    \label{tab:freecheck}
\end{table*}

\section{Qualitative Analysis}
\label{sec:qualtative_analysis}

\begin{figure*}[t]
   \begin{center}
   \includegraphics[width=1.0\textwidth]{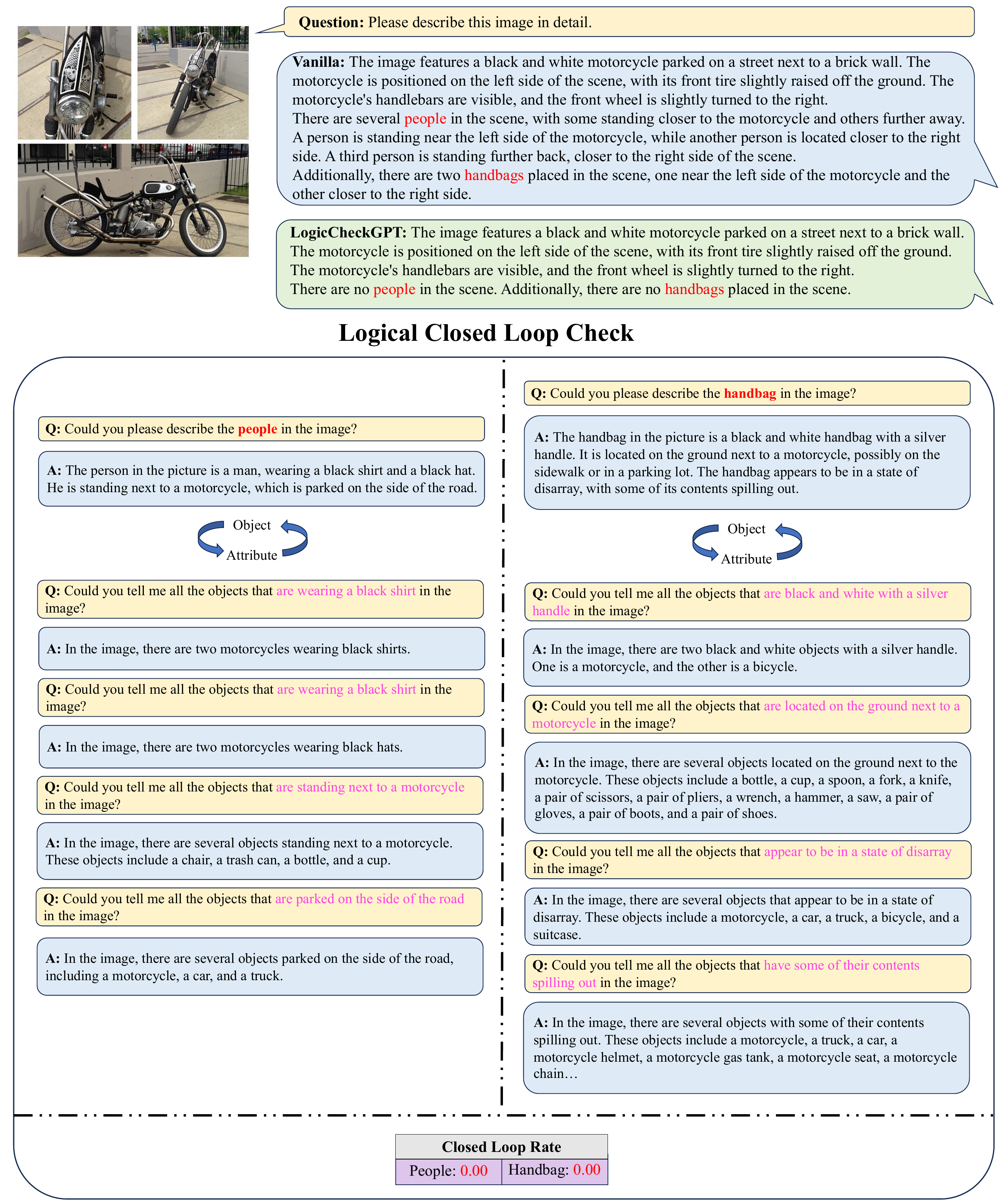}
   \end{center}
   \caption{
    The example of our LogicCheckGPT for mPLUG-Owl. The hallucinated objects are highlighted in \textcolor{red}{red} and attributes are highlighted in \textcolor{magenta}{magenta}.
   }
   \label{fig:case_mplug}
\end{figure*}

\begin{figure*}[t]
   \begin{center}
   \includegraphics[width=1.0\textwidth]{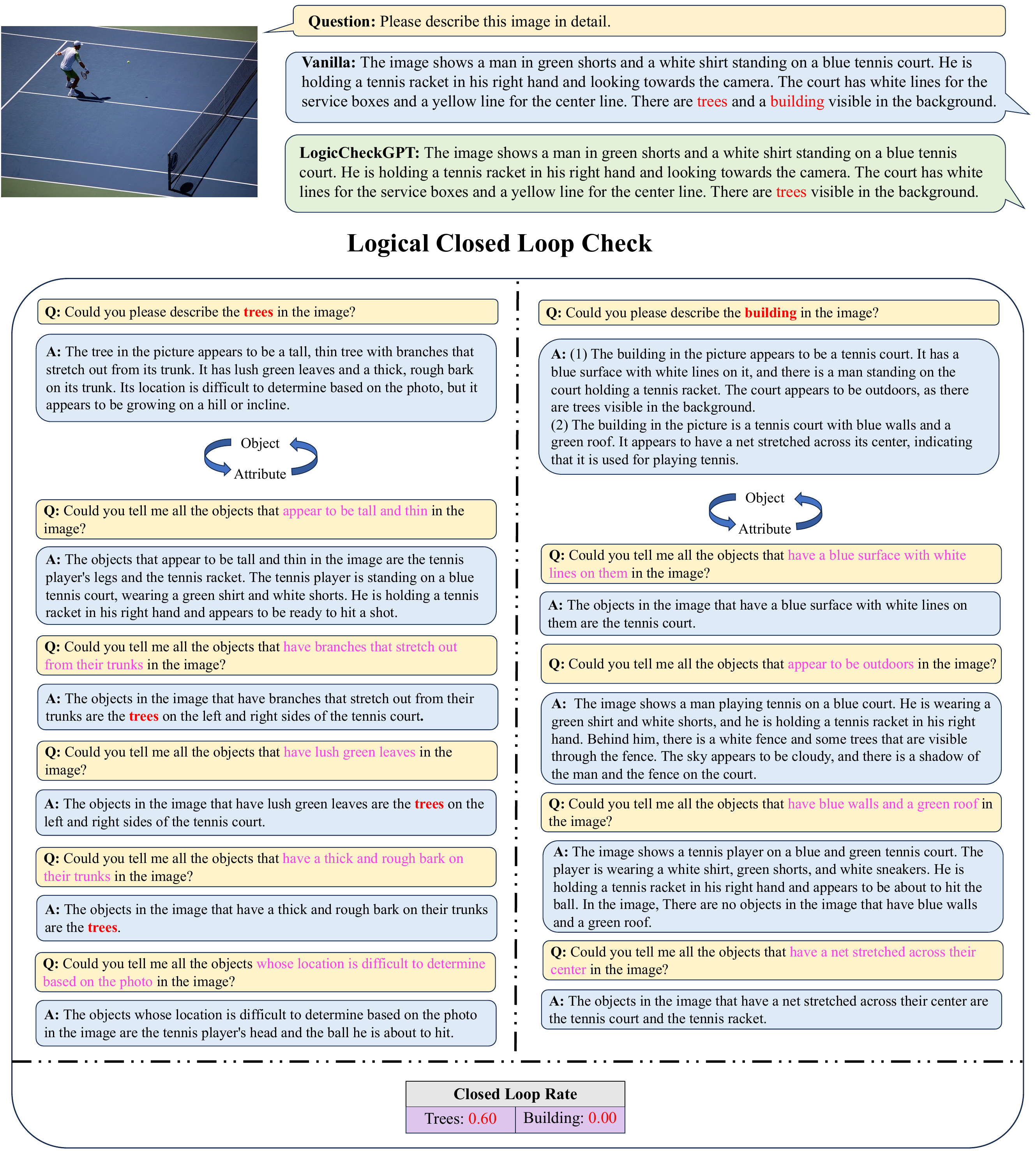}
   \end{center}
   \caption{
    The example of our LogicCheckGPT for MiniGPT-4. The examinee hallucinated objects are highlighted in \textcolor{red}{red} and attributes are highlighted in \textcolor{magenta}{magenta}. We sample two descriptions of ``Building'' to obtain sufficient attributes.
   }
   \label{fig:case_minigpt}
\end{figure*}

\begin{figure*}[t]
   \begin{center}
   \includegraphics[width=1.0\textwidth]{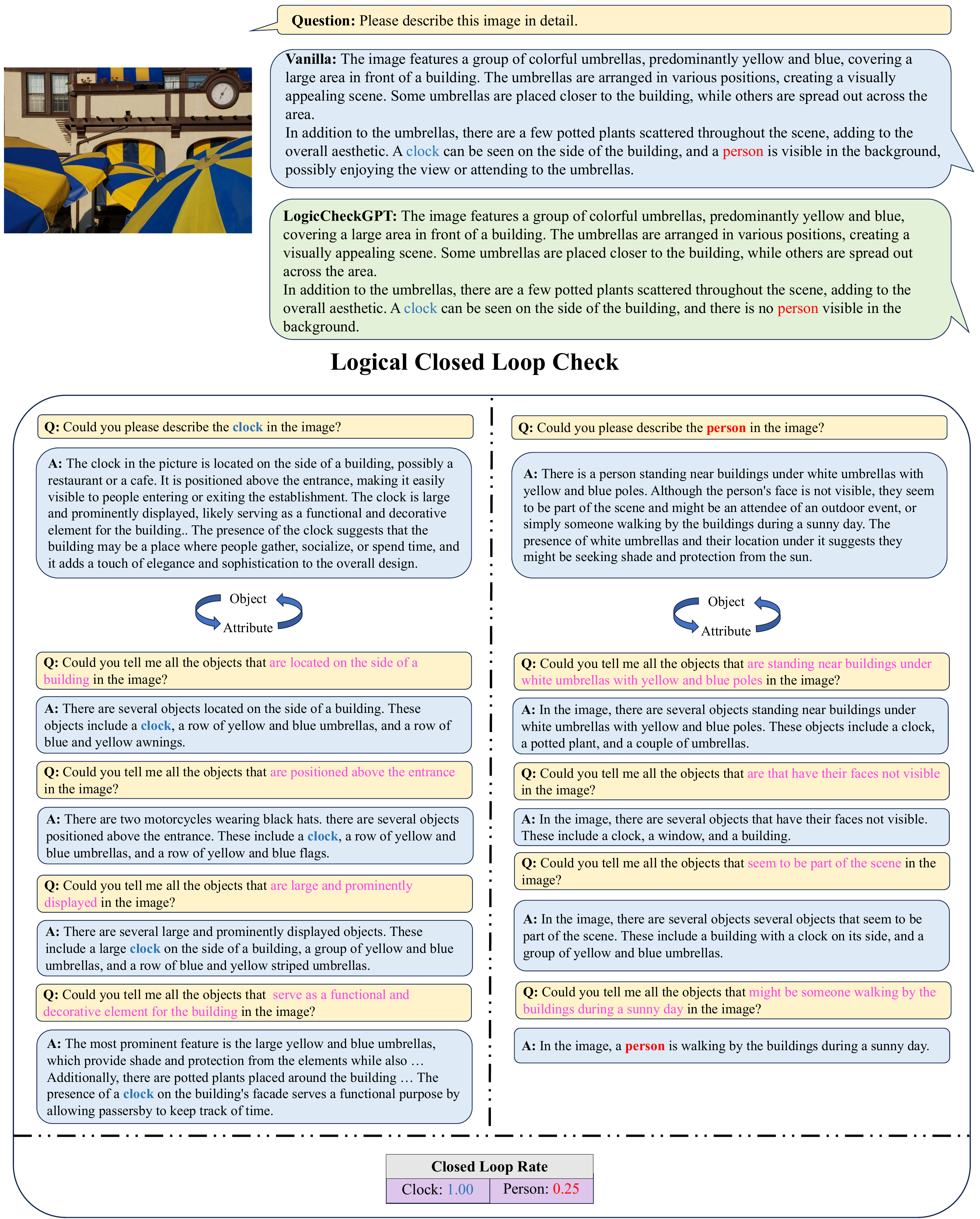}
   \end{center}
   \caption{
    The example of our LogicCheckGPT for LLaVA-1.5. The examinee hallucinated objects are highlighted in \textcolor{red}{red}, the examinee existent objects are highlighted in \textcolor{blue}{blue}, and attributes are highlighted in \textcolor{magenta}{magenta}.
   }
   \label{fig:case_llava}
\end{figure*}

In this section, we visualize several examples of our LogicCheckGPT applied to various LVLMs, as shown in Fig. \ref{fig:case_mplug}, \ref{fig:case_minigpt}, \ref{fig:case_llava}. As depicted in \ref{fig:case_mplug}, for the MPLUG-Owl example, our method effectively identifies the hallucinated objects ``people'' and ``handbag'' both of which have logical closed-loop rates of $0.0$. In contrast, for MiniGPT-4 (see Fig. \ref{fig:case_minigpt}), our method successfully detects the ``building'' but fails to identify the hallucinated object ``trees''. This failure is attributed to the fact that the described hallucinated attributes are strongly correlated with the object and may inadvertently reveal the object's identity. For instance, phrases like ``have branches that stretch out from their trunks'' and  ``have lush green leaves'' suggest that the objects are trees. In addition, for LLaVA-1.5 in Fig. \ref{fig:case_llava}, LogicCheckGPT assigns a score of $1.00$ to the existent object ``clock'' and $0.25$ to the hallucinated object ``person''. Since the score for "person" is below the threshold utilized in our study, we can detect the hallucinated object "person".

\end{document}